# RFL-CDNet: Towards Accurate Change Detection via Richer Feature Learning

Yuhang Gan [a,b,c,d,e,†], Wenjie Xuan[a,b,c,d,†], Hang Chen[a,b,c,d], Juhua Liu[a,b,c,d,*] , Bo Du[a,b,c,d]

[a] School of Computer Science, Wuhan University, Wuhan, China.

[b] National Engineering Research Center for Multimedia Software, Wuhan University, Wuhan, China.

[c] Institute of Artificial Intelligence, Wuhan University, Wuhan, China.

[d] Hubei Key Laboratory of Multimedia and Network Communication Engineering, Wuhan University, Wuhan, China.

[e] Land Satellite Remote Sensing Application Center, MNR, Beijing, China.

**Abstract.** Change Detection is a crucial but extremely challenging task of remote sensing image analysis, and much progress has been made with the rapid development of deep learning. However, most existing deep learning-based change detection methods mainly focus on intricate feature extraction and multi-scale feature fusion, while ignoring the insufficient utilization of features in the intermediate stages, thus resulting in sub-optimal results. To this end, we propose a novel framework, named RFL-CDNet, that utilizes richer feature learning to boost change detection performance. Specifically, we first introduce deep multiple supervision to enhance intermediate representations, thus unleashing the potential of backbone feature extractor at each stage. Furthermore, we design the Coarse-To-Fine Guiding (C2FG) module and the Learnable Fusion (LF) module to further improve feature learning and obtain more discriminative feature representations. The C2FG module aims to seamlessly integrate the side prediction from previous coarse-scale into the current fine-scale prediction in a coarse-to-fine manner, while LF module assumes that the contribution of each stage and each spatial location is independent, thus designing a learnable module to fuse multiple predictions. Experiments on several benchmark datasets show that our proposed RFL-CDNet achieves state-of-the-art performance on WHU cultivated land dataset and CDD dataset, and the second best performance on WHU building dataset. The source code and models are publicly available at https://github.com/Hhaizee/RFL-CDNet.

# 1    Introduction

Change Detection (CD) in remote sensing images aims to identify specific geographic element differences between bi-temporal images within the same geographic area, which takes registered bi-temporal images as the input and outputs a pixel-wise change map [1]-[3]. Due to the availability of large-scale high-resolution remote sensing images and the success of deep learning techniques, CD has become an active topic in the computer vison community and is widely used in many real-world applications, such as land-use monitoring [4], resource and environment monitoring [5], disaster assessment [6], *etc*. However, this task remains challenging due to the complex and heterogeneous appearance of geographical elements at different times.

In the past decades, a wide variety of traditional methods have been proposed for change detection, which can be roughly divided into the following categories: algebra-based methods [7], transformation-based methods [8], classification-based methods [9], and traditional machine-learning-based methods [10]. Although these traditional methods are more explainable, they mainly rely on elaborately hand-crafted features and tedious post-processing rules, which are very complicated and prone to error accumulation. Therefore, their performance is always sub-optimal and of low robustness, also very time consuming. Recently, benefiting from the rapid development of deep learning, many deep-learning-based methods have been proposed, and due to the powerful feature representation capability of Convolutional Neural Networks (CNNs), they have demonstrated superior performance and robustness over traditional methods. Deep-learning-based methods usually use CNNs-based Siamese neural networks to extract discriminative features from the input bi-temporal images, and then predict the final change results via metric-based networks [11]-[13] or Fully Convolutional Networks (FCN) [14]-[16]. Generally, to detect changes accurately, it is a prerequisite to extract discriminative features from bi-temporal images, *i.e.*, features associated with changed pixel pairs are farther apart from each other, while unchanged pixel pairs are close in the feature space. To this end, most existing deep-learning-methods focus on extracting and fusing multi-scale features for the aim of producing more accurate change maps. Despite these approaches having achieved impressive performance, they still suffer from insufficient utilization of multi-scale features, especially those in the intermediate stages.

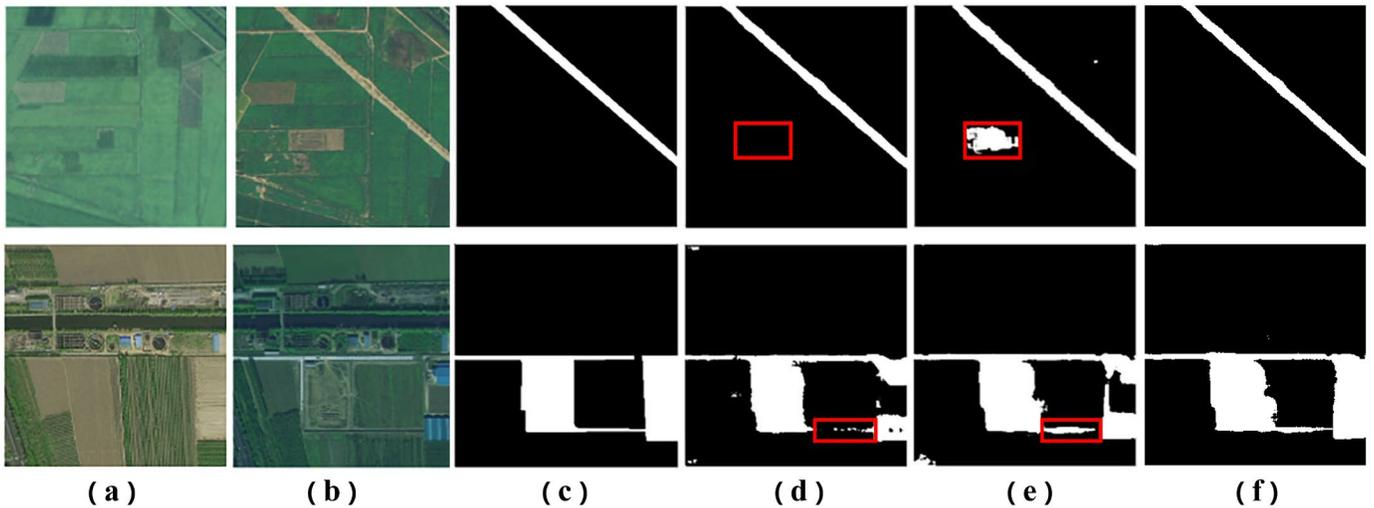

**Fig. 1** Illustration of side predictions at different stages of SNUNet. Images from left to right are (a) pre-event images, (b) post-event images, (c) ground truths, (d) side prediction at the stage 0, (e) side prediction at stage 3, and (f) the result of our proposed RFL-CDNet, respectively. We denote the stage as the down-sampling ratio of features for conducting prediction (refer to Sec.3.1). For SNUNet, stage 0 corresponds to the feature with the same size to the original image, and stage 3 has the smallest feature size.

Due to the hierarchical nature of CNNs, most existing CD methods usually design modules like Atrous Spatial Pyramid Pooling (ASPP) [17] or UNet-style networks [18][19] to extract and fuse multi-scale features, thereby obtaining the multi-scale features of the bi-temporal images. Then, these feature maps are used to produce the change map via a prediction module. For example, SNUNet [14] utilizes a Siamese Nested-UNet [19] for feature extraction and fusion, and takes the features output by Nested-UNet for prediction. During this processing, the features at intermediate stages are only used for subsequent feature fusion, and do not directly participate in the final prediction. This may have little impact on these tasks that take one single image as input, including object detection, instance segmentation, *etc*. However, change detection always takes bi-temporal images as input. As shown in Fig. 1 (a) and (b), due to significant differences in imaging conditions, the two input images themselves probably have severe domain gaps. If the features of intermediate stages only participate in feature fusion, it may lead to underutilization of features. Inspired by HED-based edge detection methods [20]-[22], which not only extract multi-scale features but also predict multiple side predictions under deep multiple supervision, we have reason to believe that existing CD methods neglect the supervision of intermediate features because CD and edge detection are both pixel-level prediction tasks that focus more on edges. Thus, there still exists room for further improving change detection performance. As shown in Fig. 1, (d) and (e) are the side predictions at stage 0 and stage 3 of SNUNet, respectively. For these two examples, regardless of which side prediction we take as the final result, there would be false positives or false negatives. If we design some strategies to take advantage of intermediate features and integrate these side predictions, it can effectively improve the final result, as shown in Fig. 1 (f).

In this paper, we propose a novel change detection framework, namely RFL-CDNet, which aims to achieve accurate change detection via learning richer multi-scale features. Following HED-based edge detection methods, we add deep multiple supervision to the intermediate stages and obtain multiple prediction results at different stages. This not only benefits multi-scale representation learning, but also incorporates more supervision for network training. Furthermore, we design a Coarse-To-Fine Guiding (C2FG) module and a Learnable Fusion (LF) module to take full advantage of richer multi-scale features and side predictions. Specifically, the C2FG module adaptively combines semantics from coarse-scale predictions with the details from fine-scale features. The C2FG module is composed of LSTM cells to encode the feature and side prediction from the adjacent coarse scale, and then fuse them with the current-scale features for the current side prediction. We illustrate this design is different from previous methods that only take features into consideration, where our C2FG is advanced in keeping history features for enhancing scale-related feature learning. For multiple side predictions from different stages, different from the typical scale-related weighted fusion strategy, we assume that different scales and positions contribute differently to the final result. Therefore, we propose an LF module to ensemble the side predictions at all stages in a more flexible manner, which is more powerful than the regular global weighting strategy. The architecture of our proposed framework is shown in Fig. 2.

The main contributions can be summarized in three folds:

- We propose a novel change detection framework, namely RFL-CDNet, which achieves richer feature learning by simultaneously combining multi-scale features with multi-stage side predictions, thereby achieving more accurate change detection;
- We design two modules, *i.e.*, C2FG and LF, where the C2FG module makes full use of multi-scale

representations in a coarse-to-fine manner by exploiting semantic cues from coarse-scale side predictions for guidance, while the LF module effectively integrates multiple scale-related predictions via a learnable fusion module. Ablation studies prove the effectiveness of both modules;

- Extensive experiments on three challenging datasets demonstrate the effectiveness and universality of our proposed RFL-CDNet. Most importantly, our RFL-CDNet sets new state-of-the-art (SOTA) performance with an *F1-score* of 72.28% on WHU cultivated land dataset, 96.12% on CDD dataset, and it achieves the second best performance with an *F1-score* of 91.39% on WHU building dataset.

The rest of the paper is organized as follows. In Section 2, we briefly review the related works of change detection and multi-scale representation learning. In Section 3, we introduce our proposed framework before analyzing each module. Section 4 reports and discusses the experimental results. Finally, we conclude in Section 5.

## 2 Related Works

*2.1 Traditional CD Methods*

As aforementioned, traditional CD methods can be roughly classified into four categories: 1) **Algebra-based methods** involve performing channel-wise algebraic operations directly on the registered bi-temporal images. These operations include image difference, regression, and change vector analysis [7]. However, finding a suitable threshold to distinguish changed pixels from unchanged ones is both scene-dependent and time-consuming. 2) **Transformation-based methods** aim to convert bi-temporal images into specific feature spaces, which helps reduce the pixel differences of unchanged regions and highlight the changed information. However, methods such as [8] struggle to handle high-resolution images due to their reliance on empirically designed features. 3) **Classification-based methods** like [9] identify change regions by comparing the pre-generated land cover labels of geographic elements between bi-temporal images. However, the classification errors in the pre-generated labels would accumulate in the final change maps and inevitably reduce the accuracy of predicted results. 4) **Machine learning-based methods** utilize traditional machine learning algorithms, such as random forest regression and decision tree [10], to determine whether the specified area has changed. Although most of these traditional methods are simple and explainable, they show poor robustness in real-world scenarios as their performance is prone to noise and limited by handcrafted features.

*2.2 Deep-learning-based CD Methods*

In recent years, many deep-learning-based CD methods have been proposed and reveals better performance than traditional methods. Existing deep-learning-based CD frameworks usually include the following two steps: first, feature extraction modules extract features with significant change information from bi-temporal images, and then a prediction module produces change results based on the extracted features. Recent works focus on improving CD methods mainly from two perspectives. On the one hand, discriminative features are of great significance for relieving pseudo changes caused by different imaging conditions. Therefore, many attempts explore more discriminative features by aggregating multi-scale information or designing more intricate structures. Earlier methods like FC-EF, FC-Siam-conc, and FC-Siam-diff [15] utilized UNet [18] as the backbone network for multi-scale feature extraction. They investigated three distinct feature fusion strategies, resulting in significant improvements in both performance and processing speed compared to their predecessors. Chen et al. [11] introduced the influential STANet, which incorporated a spatial-temporal self-attention module to the Siamese network for investigating spatial-temporal dependencies between bi-temporal images. Fang et al. [14] proposed a densely connected SNUNet, achieving state-of-the-art performance on the CDD dataset. SNUNet adopted a Siamese Nested-UNet [19] architecture as the feature extractor and utilized an Ensemble Channel Attention Module (ECAM) to aggregate multi-scale features, thereby enhancing discriminative feature representations. $S^2$PNet [23] designed a parallel spatial-channel attention mechanism and a self-structured feature pyramid to refine features of different levels. Recently, following the success of Transformer in the fields of Natural Language Processing (NLP) and Computer Vision (CV), Transformer has also been used for the feature extraction of bi-temporal images. The most representative method is Bi-temporal Image Transformer (BIT) [16] and Visual Change Transformer (VcT) [24], which efficiently and effectively model the spatial–temporal contexts for more discrimination features. Wang et al. [25], Li et al. [26], and Ke et al. [27] further combines transformers with CNNs to learn stronger feature representations for improvements.

On the other hand, the strategy adopted in the prediction module to determine whether pixels have changed according to the extracted features also plays a crucial role in CD. Existing methods can be roughly divided into metric-based and FCN-based methods. Metric-based methods [11]-[13] determine the change by comparing feature distance between bi-temporal images, whereas FCN-based methods [14]-[16] employ Fully Convolutional Networks (FCN) to conduct pixel-level predictions according to the features of bi-temporal images. Our proposed RFL-CDNet is built on SNUNet, which employs a Siamese Nested-UNet for feature extraction and utilizes an FCN to predict change maps.

*2.3 Multi-Scale Representation Learning*

Multi-scale representation learning plays a crucial role in enhancing feature representations for most computer vision tasks, such as edge detection [20]-[22], object detection [28], semantic segmentation [17], *etc*. The core idea is to integrate multi-scale features captured from various receptive fields, like the hierarchical features from the CNN-based backbone networks, to improve feature representations. Such multi-scale representations could provide rich context, largely enhancing the capability of networks on downstream tasks. For instance, FPN [29] introduces a feature pyramid network to integrate multi-scale features from different stages of VGG or ResNet with few extra computations, benefiting object detection tasks [28]. In semantic segmentation, Deeplab [17] employs Atrous Spatial Pyramid Pooling (ASPP) to expand the receptive field, which captures richer context information with few extra parameters. Recently, DenseNet [30], HRNet [31], and UNet [18] resort to skip connections to aggregate features from adjacent scales or between encoders and decoders, having achieved impressive performance in classification, pose estimation, and medical applications. Res2Net [32] further constructs hierarchical residual-like connections within one single residual block to represent multi-scale features at a granular level, increasing the receptive fields for each network layer. Moreover, Yan et al. [33][34] proves that the powerful transformer still benefits from multi-scale representation learning in change detection.

Considering that changed regions have various sizes, multi-scale representations are essential in capturing object features and change features to detect changes. Previous change detection methods resorted to UNet [18] or its variations like Nested-UNet [19] and T-UNet [35] to learn multi-scale representations. In contrast, this paper mainly focuses on improving change detection with richer representations from the intermediate stages. We propose a novel C2FG module for multi-scale feature fusion in a coarse-to-fine manner, which simultaneously integrates information from multi-scale features and multi-stage side predictions, thus exploiting richer features for change detection. It reveals advantages in keeping long-term feature memory to adaptively incorporate fine-scale details to coarse-scale semantics.

## 3 Methodology

In this section, we first introduce the overview architecture of our RFL-CDNet, and then illustrate details of three main components, including the backbone network and our designed modules for richer representation learning, *i.e.*, the Coarse-To-Fine Guiding (C2FG) module and the Learnable Fusion (LF) module. Finally, we define the full loss function.

### *3.1 Overview Architecture*

Given a pair of registered bi-temporal remote sensing images in the same geographic area, *i.e.*, the pre-event image $X^A = \{x_{ij.}^A\}_{H \times W \times 3}$, and the post-event image $X^B = \{x_{ij.}^B\}_{H \times W \times 3}$, change detection method aims to accurately predict a binary change map $\hat{Y} = \{\hat{y}_{ij}\}_{H \times W}$, $\hat{y}_{ij} \in \{0, 1\}$, where 1 represents changed pixel and 0 means no change. The ground-truth change map is denoted as $Y = \{y_{ij}\}_{H \times W}$, $y_{ij} \in \{0, 1\}$. For convenience, we define the stage $s$ as the down-sampling ratio of the feature map $F^s = \{f_{ij.}\}_{H' \times W' \times C'}$, where $s = log_2(H/H')$. Our work is built upon SNUNet [15] with four stages, *i.e.*, $s = 0, 1, 2, 3$.

Fig. 2 depicts the overall architecture of our proposed RFL-CDNet framework, which consists of three components: 1) a backbone built upon Nested-UNet for feature extraction, 2) a C2FG module exploiting richer intermediate features for multi-scale representation learning, and 3) an LF module to ensemble multi-stage side predictions. First, the backbone extracts representations of land objects and change features from bi-temporal images. We introduce a deep multiple supervision scheme to the backbone for boosting intermediate features. Specifically, we append the ECAM of SNUNet to each stage and produce the side prediction $\hat{Y}^s$ based on the intermediate features $F^s$. $\hat{Y}^s$ is supervised by the ground truth $Y$. Then, the C2FG module exploits rich intermediate features via progressively integrating the current scale feature $F^s$ and side prediction $\hat{Y}^{s+1}$ through LSTM cells in a coarse-to-fine manner. Thus, the C2FG module can adaptively aggregate semantic cues from coarse-scale side predictions with location cues from fine-scale features. Finally, we introduce an LF module to fuse all side predictions through the learned weight map $W^s$, generating a more robust change map $\hat{Y}$. The key functions of these three components are elaborated in detail as follows.

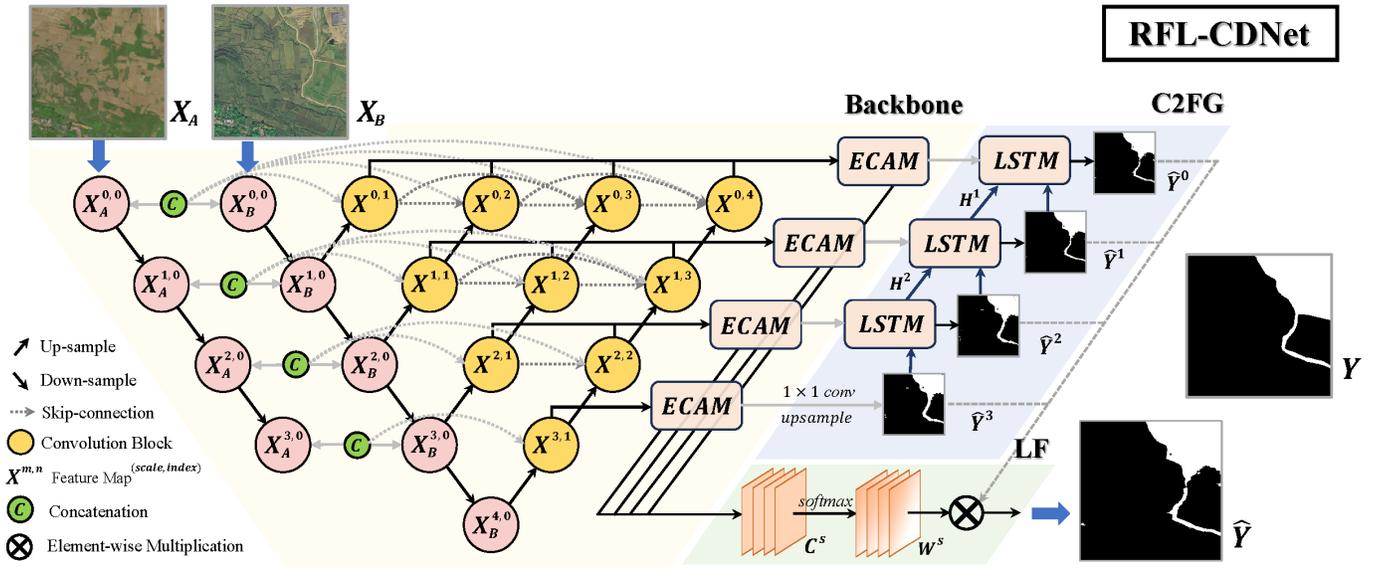

**Fig. 2** The overall architecture of our proposed RFL-CDNet. It comprises of: 1) A Nested-UNet with deep multiple supervision to facilitate intermediate feature learning, 2) A C2FG module to exploit rich intermediate features for multi-scale representation learning, 3) A LF module to ensemble side predictions at all stages for robust change detection. Better viewed in color.

*3.2 Backbone*

Following SNUNet, our RFL-CDNet adopts a Siamese Nested-UNet to extract hierarchical features $X^{m,n}$, where *m* and *n* represent the stage and feature index, respectively, as shown in Fig. 2. Such Siamese densely-connected backbone has two main advantages: 1) the Siamese encoder with shared parameters can perceive change features from registered bi-temporal images, and 2) the encoder-to-decoder and decoder-to-decoder dense connections compensate for the information loss caused by successive down-samplings. SNUNet proposed an ECAM to aggregate features from different semantic levels, suppressing semantic gaps and localization differences among multi-scale features. Then, SNUNet outputs the final change map based on the aggregated features. Different from SNUNet, we append ECAM to all stages to get richer multi-scale features inspired by RCF [21]. And we employ deep multiple supervision scheme to obtain more discriminative intermediate representations from sub-decoders, facilitating feature learning at intermediate stages. Specifically, with the features refined by ECAM, we concatenate them and append a $1 \times 1$ convolutional layer to predict the change map, termed side prediction $\hat{Y}^s$. All side predictions are supervised by the ground-truth. Since the intermediate features are updated by the loss computation of side predictions rather than merely the final change map, it not only strengthens scale-related feature learning but also helps quicker convergence of the network.

*3.3 Coarse-To-Fine Guiding (C2FG) Module*

Considering the increment brought by deep multiple supervision in Sec. 4.4.1, it implies insufficient feature learning despite the densely connected Nested-UNet backbone. This is attributed to the gradient vanishing among the long-term iterative feature fusion. Inspired by the success of the coarse-to-fine optimization in field of Generative Adversarial Nets (GANs) [36] and diffusion model [37], we design a novel Coarse-to-Fine Guiding (C2FG) module to exploit richer intermediate features for multi-scale representation learning, further releasing the capability of the network in change detection. Generally, coarse-scale features from deeper stages contain semantic cues, while fine-scale features from shallow stages maintain precise location and details. Therefore, as shown in Fig. 2, C2FG module is constructed by a series of LSTM cells, which aggregate intermediate features in a coarse-to-fine manner to progressively bridge the semantic gap. We start from conducting predictions via the coarsest but semantic-high features to capture the interested changes roughly, and then gradually refines boundaries by incorporating location cues and details from fine-scale features. While the widely-used FPN [29] provides a solution for aggregating multi-scale features by simple feature concatenation and convolution, we notice it suffers from insufficient utilization of multi-scale features, *i.e.*, the impact of early-fused features grows weak during the long iterative feature fusion process. Thus, we choose LSTM for its long-term memory of historical multi-scale features. LSTM stores historical features in its hidden states for memory, which helps to keep semantic cues from coarse-scale features in our design. Under the guidance of these semantic cues, the model is encouraged to focus on learning scale-specific features and refine representations progressively.

Specifically, each LSTM cell contains a hidden state $H^s$ for long-term feature memory and three gates. Given the input features $F^s$ and the previous hidden state $H^{s+1}$, it utilizes the input gate $G_i$, forget gate $G_f$ and output gate $G_o$ to mine useful information step by step. The gated signal $g_k$ can be computed as follows:

$$g_k = \sigma(W_k * [H^{s+1}, F^s] + b_k), \qquad (1)$$

where $W_k, b_k, k = i, f, o$ are convolutional weights and bias, and $\sigma(\cdot)$ represents the sigmoid function. And the side prediction $\hat{Y}^s$ can be defined as follows:

$$\hat{Y}^s = g_f * \hat{Y}^{s+1} + g_i * tanh(W_c * [H^{s+1}, F^s] + b_c), \qquad (2)$$

where $W_c$ and $b_c$ are convolutional weights and biases. Referring to historical features kept in the hidden state $H^{s+1}$ and coarse-scale prediction $\hat{Y}^{s+1}$, the side prediction $\hat{Y}^s$ can exploit historical information and relieve redundant feature extraction in $F^s$. Therefore, our C2FG module also helps emphasize scale-related features. Then, we update the hidden state $H^s$ as follows:

$$H^s = g_o * tanh(\hat{Y}^s), \qquad (3)$$

where $H^s$ adaptively reserves multi-scale features through the side prediction $\hat{Y}^s$ and output gate $G_o$. Noted that we initialize the hidden state with the side prediction for the coarsest stage $s = 3$. For other stages, the side prediction and the hidden state are computed as Eq.1 and Eq.2. Thus, the proposed RFL-CDNet can be treated as simultaneously integrating information from multi-scale features and multi-stage outputs, allowing for learning richer features for change detection.

### 3.4 Learnable Fusion (LF) Module

Given the side predictions produced by the nested sub-detectors in the backbone, it is intuitive to build a stronger and more robust change detector through ensemble learning. However, as the sub-detectors are highly related due to some shared features, the regular global fusion strategy, *i.e.*, computing the average or weighted sum of various predictions, would cause sub-optimal results. Another reason is the side predictions have different confidence varied by objects and scales, and probably contain false positives or false negatives due to the semantic gap as shown in Fig. 1. In fact, coarse-scale features have large receptive fields. Therefore, such features are sensitive to the changes of large-scale geographic elements but fail to maintain delicate boundary structures. In contrast, fine-scale features align with the boundary of change regions well but contain many false positives caused by the lack of global semantics. Therefore, only adopting scale-related global weights for prediction ensembling is unsuitable.

Here we assume predictions at each spatial location and each scale are independent, and propose a Learnable Fusion (LF) module to adaptively fuse multi-scale predictions at pixel level. As shown in Fig. 2, LF aims to learn scale-related pixel-wise confidence maps $C^s = \{c_{ij}^s\}_{H \times W}$, describing how confident the side prediction is at position $(i,j)$. Specifically, we utilize $1 \times 1$ convolution to further extract scale-related confidence information $C^s$ from $F_s^{'}$ and compute pixel-level weights $W^s = \{w_{ij}^s\}_{H \times W}$ via softmax, which can be formulated as follows:

$$(W^1, W^2, \dots, W^s) = softmax(cat(C^1, C^2, \dots, C^s)), \tag{4}$$

where $cat(\cdot)$ represents concatenation. Then, LF conducts element-wise multiplication of side predictions with scale-related pixel-level weights $W^s$ to ensemble multi-scale predictions delicately. Then, the final change map $\hat{Y}$ can be computed as follows:

$$\hat{Y} = \sum_s W^s \otimes \hat{Y}^s, \tag{5}$$

where $\otimes$ denotes element-wise multiplication.

*3.4 Loss Function*

Referring to SNUNet [14], we adopt a hybrid loss function of weighted cross-entropy loss $L_{wce}$ and dice loss $L_{dice}$ to relieve the impact of class imbalance caused by the predominance of unchanged pixels. For the final prediction $\hat{Y}$, the loss is computed as follows:

$$L_{wce}(\hat{Y}, Y) = -w_- \cdot \sum_{y_{ij}=1} log(\hat{y}_{ij}) - w_+ \cdot \sum_{y_{ij}=0} log(1 - \hat{y}_{ij}) \tag{6}$$

$$L_{dice}(\hat{Y}, Y) = 1 - \frac{2 \cdot |Y \cap \hat{Y}|}{|Y| + |\hat{Y}|}, \tag{7}$$

$$L(\hat{Y},Y) = L_{wce}(\hat{Y},Y) + L_{dice}(\hat{Y},Y), \tag{8}$$

where $w_+ = \frac{\sum_{i,j} \mathbb{I}(y_{ij}=1)}{H \times W}$ and $w_- = \frac{\sum_{i,j} \mathbb{I}(y_{ij}=0)}{H \times W}$ represent the class-balanced weights for unchanged and changed pixels respectively. To conduct deep multiple supervision, we also compute the hybrid loss for all side predictions. Therefore, the overall loss L for our proposed RFL-CDNet is computed as,

$$L = \sum_s \lambda_s * L(\hat{Y}^s, Y) + \mu * L(\hat{Y}, Y), \tag{9}$$

where $\lambda_s$ ($s = 0,1,2,3$) and $\mu$ are coefficients for side predictions and the final result, respectively. In our experiments, we empirically set all side weights $\lambda_s$ to 0.25 and $\mu$ to 1.0.

## 4 Experiments

In this section, we implement our proposed RFL-CDNet framework based on SNUNet (2021) and evaluate its effectiveness on three benchmark datasets, including WHU cultivated land, CDD and WHU building datasets. We compare RFL-CDNet with the baseline as well as other previous state-of-the-art methods, and conduct extensive ablation studies to carefully analyze the effectiveness of each proposed module.

### 4.1 Datasets and Evaluation Metrics

**WHU Cultivated Land Dataset** [38] was collected, labeled, and constructed by the State Key Laboratory of Information Engineering in Surveying, Mapping and Remote Sensing of Wuhan University. It mainly focuses on the change of cultivated land and the image resolution is 1~2 meters. This dataset provides 3,194 bi-temporal images for training and 1,000 bi-temporal images for test, and the image size is 512×512. However, since this dataset is mainly used for public change detection competitions, the ground-truth of test set is not pubic, we randomly split the training set, and take 500 bi-temporal images with ground-truth as our test set, and the rest as the training set.

**CDD** [39] is a widely-used dataset for change detection in high resolution remote sensing imagery, which contains 11 pairs of bi-temporal images obtained from Google Earth in different seasons with resolution from 0.03 to 1 meter. The original dataset consists of seven 4,725×2,700 bi-temporal images and four 1,900×1,000 bi-temporal images. Since the image size is too large to be processed directly, the original images are cropped to image patch with the size of 256×256, thus generating 10,000 bi-temporal images for training, 3,000 bi-temporal images for validation, and 3,000 bi-temporal images for test.

**WHU Building Dataset** [40] focuses on the change of buildings, and contains a pair of very high-resolution remote sensing imagery of 32,507×15,354 pixels with a spatial resolution of 0.075m. The original image is divided into a training image of 21,243×15,354 pixels and a test image of 11,265×15,354 pixels. Due to the memory constraints, we further crop the images into 256×256-pixel patches without overlap, resulting in 4,980 bi-temporal images for training and 2,700 bi-temporal images for test.

We follow the classic evaluation protocols and adopt Precision ($P$), Recall ($R$), and *F1-score* as metrics to evaluate the performance of change detection. *Precision* represents the accuracy of detection results, while *Recall* reflects the completeness of the predicted changed regions. Since there is a trade-off between *Precision* and *Recall*, we adopt *F1-score* as the evaluation criterion in this paper. In real-world applications such as environment monitoring and disaster assessment, *Recall* is often more concerned by users. This is because change detection is utilized to locate possibly changed regions as candidate regions for further checks like identifying the exact category of geographical element changes. However, for comprehensively evaluating the detection performance, we adopt *F1-score* in our experiments. The higher *F1-score* reflects the better detection performance.

*4.2 Implementation Details*

We implement our proposed RFL-CDNet framework based on SNUNet. Following SNUNet, Adam was adopted for model optimization. The weight decay was set to 0.0001, the momentum to 0.9, and the batch size to 8. The initial learning rate was set to 0.001 and decayed by 0.5 every 8 epochs. The weights of each convolutional layer are initialized by the Kaiming normalization. All experiments were conducted on the high-performance servers with NVIDIA Tesla V100 (16G) GPUs. The model was trained with 4 GPUs for 100 epochs to make the model converge, and evaluated with 1 GPU. To enhance the robustness of the model, we also employed some commonly used data augmentation strategies, such as random flips, random rotation, random added noise, and normalization.

## 4.3 Comparison with State-of-the-Art Methods

### 4.3.1 Comparison Methods

To demonstrate the superiority of our RFL-CDNet framework, we compared the performance of RFL-CDNet with several representative and several previous state-of-the-art change detection methods.

**FC-EF, FC-Siam-conc** and **FC-Siam-diff** [15] are the most classic FCN-based methods, which integrated UNet into the Siamese network for change detection. These three methods explored three different fusion strategies including bi-temporal images fusion, multi-level feature concatenation, and multi-level feature difference, which are the main strategies for bi-temporal feature fusion in existing change detection methods.

**STANet** [11] designs a self-attention mechanism to model spatial-temporal relationships at various scales, thereby extracting more discriminative features for change detection. It well mitigates misdetections caused by misregistration of bi-temporal images and is robust to color and scale variations.

**DASNet** [12] proposes a deep metric-learning-based change detection method, which uses dual attention modules and designs the weighted double-margin contrastive loss to improve the capacity of feature learning, thereby being more robust in distinguishing changes.

**SNUNet** [14] proposes a densely connected Siamese network, which combines Siamese network with Nested-UNet to alleviate the loss of localization information in the deep layers. This model further designs an Ensemble Channel Attention Module (ECAM) to suppress semantic gaps and localization differences among multi-scale features. Our RFL-CDNet is built upon this model.

**BIT** [16] considers that high-level concepts of the interested changes can be represented by a few semantic tokens. Based on this assumption, the Bi-temporal Image Transformer (BIT) is designed to model long-range contexts within the bi-temporal images in an efficient and effective manner.

**FTN** [33] proposes a fully transformer network to improve feature representations for change detection. It takes advantages of the transformer in modeling long-range dependency and introduces a pyramid structure to aggregate multi-scale features from the transformer for feature enhancement.

**VcT** [24] employs a visual change transformer framework to capture intra-image and inter-image cues for change detection. A token selection module is designed for mining unchanged background context information, which facilitates consistent representations of the bi-temporal images.

Since previous methods did not conduct experiments on WHU cultivated land and CDD datasets, and there is also no pre-defined data split on WHU building dataset, we reproduced all the above methods on three datasets with the same dataset settings, i.e., data splits, for fair comparisons.

*4.3.2 Performance on WHU Cultivated Land Dataset*

As mentioned above, the image resolution of WHU cultivated land dataset is 1~2 meters, so we first use it to evaluate the effectiveness of our RFL-CDNet in medium resolution remote sensing imagery. Since the comparison methods did not report their performance on WHU cultivated land dataset, we followed their setting and conducted experiments using the same training and test sets. Table 1 lists the results of our RFL-CDNet framework compared to some previous methods. RFL-CDNet achieves the best overall performance (F1-score: 72.28%) and surpasses the current best method, *i.e.*, BIT, by a relatively large margin of 1.8% of *F1-score*. While FTN achieves the best precision of 75.54%, our RFL-CDNet shows much higher *Recall* and *F1-score*. Our RFL-CDNet exhibits higher accuracy on detecting detailed changes compared with recent transformer-based methods including BIT, FTN and VcT. Moreover, compared to the baseline of our framework, *i.e.*, SNUNet, RFL-CDNet consistently improves in all evaluation metrics (*Precision*: 70.87% *v.s.* 68.88%, *Recall*: 73.75% *v.s.* 72.09%, and *F1-score*: 72.28% *v.s.* 70.45%).

Fig. 3 shows some visualization comparison examples among our RFL-CDNet and previous methods. For better illustration, we highlight these areas with red boxes, where RFL-CDNet handles well but other methods fail. As shown in Fig. 3, our RFL-CDNet achieves impressive detection results, especially at the edges of some changed regions, such as the examples in the first, second, third and fifth columns. Moreover, since our RFL-CDNet can effectively improve the capacity of multi-scale representation learning, it effectively suppresses false positives compared to other methods, as shown in the examples in the fourth and sixth columns.

**Table 1.** Performance comparison with the state-of-the-art methods on WHU cultivated land dataset. The best results are marked in **bold**, while the second-best results are underlined.

| Methods | P (%) | R (%) | F1-score (%) |
|---|---|---|---|
| FC-EF | 60.29 | 62.98 | 61.61 |
| FC-Siam-conc | 62.51 | 65.24 | 63.85 |
| FC-Siam-diff | 64.81 | 56.42 | 60.33 |
| STANet | 62.75 | 69.47 | 65.94 |
| DASNet | 60.58 | **77.00** | 67.81 |
| SNUNet | 68.88 | 72.09 | 70.45 |
| BIT | 70.84 | 70.11 | <u>70.48</u> |
| FTN | **75.54** | 63.20 | 68.82 |
| VcT | 65.90 | 66.74 | 66.32 |
| Ours | <u>70.87</u> | <u>73.75</u> | **72.28** |

*4.3.3 Performance on CDD Dataset*

We next evaluate the effectiveness of our RFL-CDNet in high-resolution remote sensing imagery on the widely-used CDD dataset with the resolution of 0.03 to 1 meter. The results of comparison with previous methods are shown in Table 2. Similarly, RFL-CDNet achieves the best performance among all evaluation metrics. Specifically, RFL-CDNet sets a new state-of-the-art result with *Precision* of 96.09%, *Recall* of 96.16%, and *F1-score* of 96.12% on CDD dataset, and outperforms the latest state-of-the-art method, *i.e.*, BIT, by the gain of 1.23% *Precision*, 0.84% *Recall*, and 1.03% *F1-score*. Benefiting from powerful multi-scale representations, our method exhibits to be superior to transformer-based change detection methods including BIT, FTN, and VcT in detecting detailed changes. Regarding the performance improvement of RFL-CDNet over the baseline SNUNet, the improvement of RFL-CDNet on CDD dataset is much superior than that of WHU cultivated land dataset among all evaluation metrics (*Precision*: 3.99% *v.s.* 1.99%, *Recall*: 6.03% *v.s.* 1.66%, and *F1-score*: 4.87% *v.s.* 1.83%).

Some examples from CDD dataset are present in Fig. 4. As visualized, our RFL-CDNet has significantly improved the performance on edges and small change areas, especially on lines with delicate structures. Thus, our RFL-CDNet shows higher precision than other methods, which benefits from rich intermediate features and advanced multi-scale representation learning through the proposed C2FG.

**Table 2** Performance comparison with the state-of-the-art methods on CDD dataset. The best results are marked in **bold**, while the second-best results are underlined.

| Methods | P (%) | R (%) | F1-score (%) |
|---|---|---|---|
| FC-EF | 84.68 | 65.13 | 73.63 |
| FC-Siam-conc | 88.81 | 62.20 | 73.16 |
| FC-Siam-diff | 87.57 | 66.69 | 75.72 |
| STANet | 83.17 | 92.76 | 87.70 |
| DASNet | 93.28 | 89.91 | 91.57 |
| SNUNet | 92.40 | 90.13 | 91.25 |
| BIT | <u>94.86</u> | <u>95.32</u> | <u>95.09</u> |
| FTN | 91.77 | 89.56 | 90.65 |
| VcT | 94.21 | 92.56 | 93.38 |
| Ours | **96.09** | **96.16** | **96.12** |

*4.3.4 Performance on WHU Building Dataset*

We also validate the effectiveness of our RFL-CDNet in very high-resolution remote sensing imagery on WHU building dataset with a resolution of 0.075m. Being different from CDD which is a category-agnostic change detection, WHU building dataset focuses on building changes. Therefore, compared with CDD dataset, although WHU building dataset has higher resolution, it is still considered more challenging. The comparison results of our RFL-CDNet with previous works are presented in Table 3. Our RFL-CDNet achieves the second highest results with *Precision* of 91.33%, *Recall* of 91.46%, and *F1-score* of 91.39% on WHU building dataset, which obtains the highest *Recall* among all compared methods. The FTN only surpasses our method by 0.82% on *F1-score*, which benefits from the pre-trained transformer backbone on ImageNet22k. Our method outperforms the baseline SNUNet by a relatively large margin of 2.17% of *F1-score*. Moreover, comparing the performance of RFL-CDNet on the WHU cultivated land with that on WHU building datasets (F1-score: 91.39% *v.s.* 72.28%) under the category-specific change detection scenario, the performance on the very high-resolution dataset, i.e., WHU building dataset is far superior to that of the medium resolution dataset, i.e., WHU cultivated land dataset. This proves that the image resolution has great influence on change detection performance.

Some examples on WHU Building dataset are illustrated in Fig.5. Similar to the performance on WHU cultivated land and CDD datasets, our RFL-CDNet outperforms previous methods in detecting the edges of change areas, *i.e.*, the buildings, and suppresses the false positives, further validating the effectiveness of the proposed method.

**Table 3** Performance comparison with the state-of-the-art methods on WHU building dataset. The best results are marked in **bold**, while the second-best results are underlined.

| Methods | P (%) | R (%) | F1-score (%) |
| --- | --- | --- | --- |
| FC-EF | 80.75 | 67.29 | 73.40 |
| FC-Siam-conc | 54.20 | 81.34 | 65.05 |
| FC-Siam-diff | 48.84 | 88.96 | 63.06 |
| STANet | 77.40 | 90.30 | 83.35 |
| DASNet | 83.77 | 91.02 | 87.24 |
| SNUNet | 91.28 | 87.25 | 89.22 |
| BIT | 91.56 | 87.84 | 89.66 |
| FTN | **94.73** | 89.83 | **92.21** |
| VcT | 93.24 | 76.58 | 84.09 |
| Ours | 91.33 | **91.46** | 91.39 |

Therefore, according to the above experimental results on WHU cultivated land, CDD, and WHU building datasets, our proposed RFL-CDNet can be applied to both category-agnostic and category-specific change detection tasks, as well as to different image resolutions, which proves the effectiveness and robustness of RFL-CDNet. Benefiting from the deep multiple supervision scheme, the coarse-to-fine guidance for richer multi-scale representation learning, and scale-spatial-related prediction ensembling strategy, our RFL-CDNet exhibits superior performance on detecting change regions, especially those with small areas and delicate boundaries. Furthermore, our RFL-CDNet sets new state-of-the-art (SOTA) performance for WHU cultivated land dataset and CDD dataset and achieves the second highest F1-score on WHU building datasets. Our method reveals competitive or better performance compared with transformer-based methods.

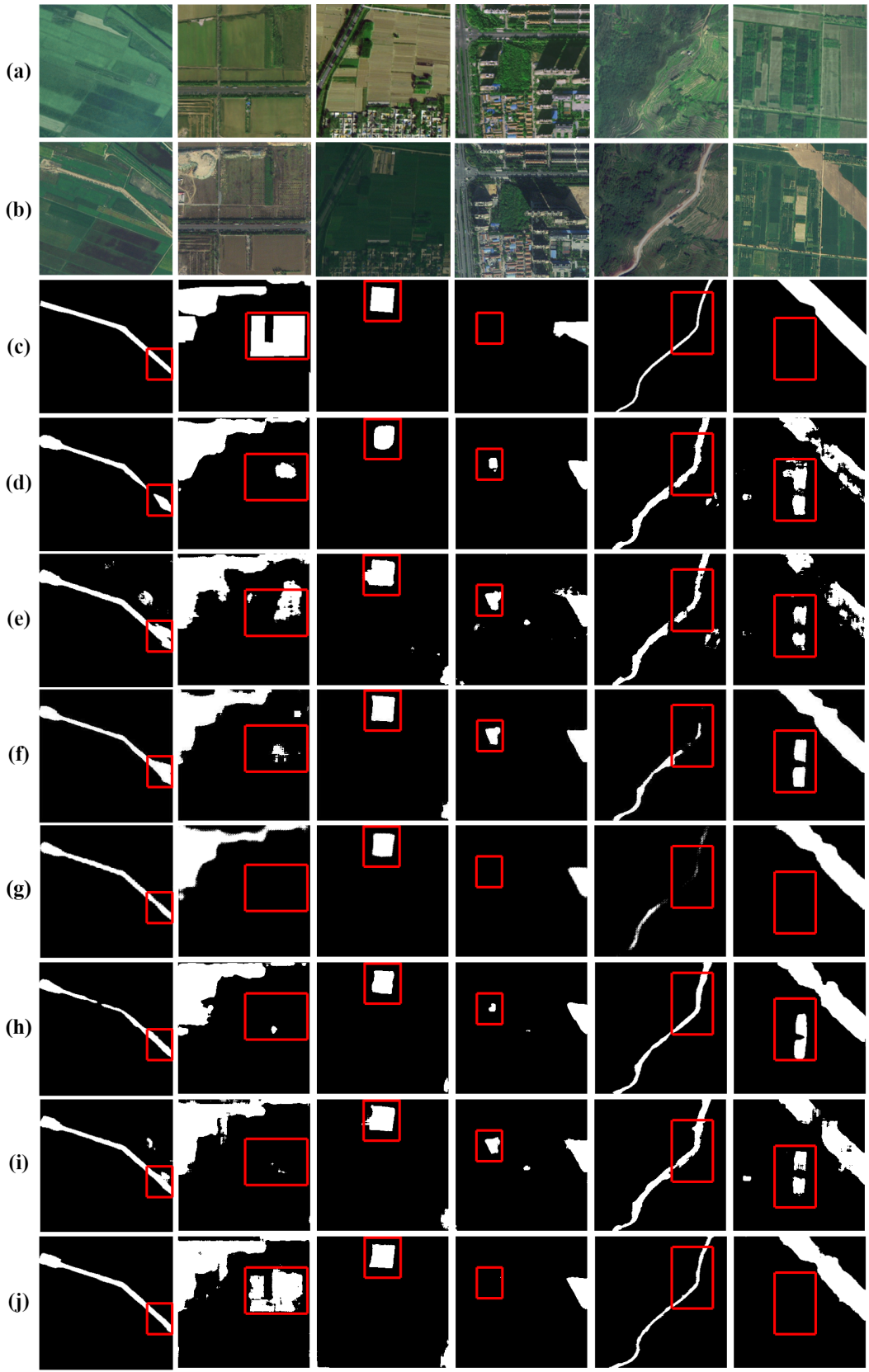

Fig. 3 Visualized comparison results on WHU cultivated land dataset. Images from top to bottom are (a) pre-event images, (b) post-event images, (c) ground truth, and the results of (d) STANet, (e) DASNet, (f) BIT, (g) FTN, (h) VcT, (i) SNUNet, and (j) our RFL-CDNet, respectively.

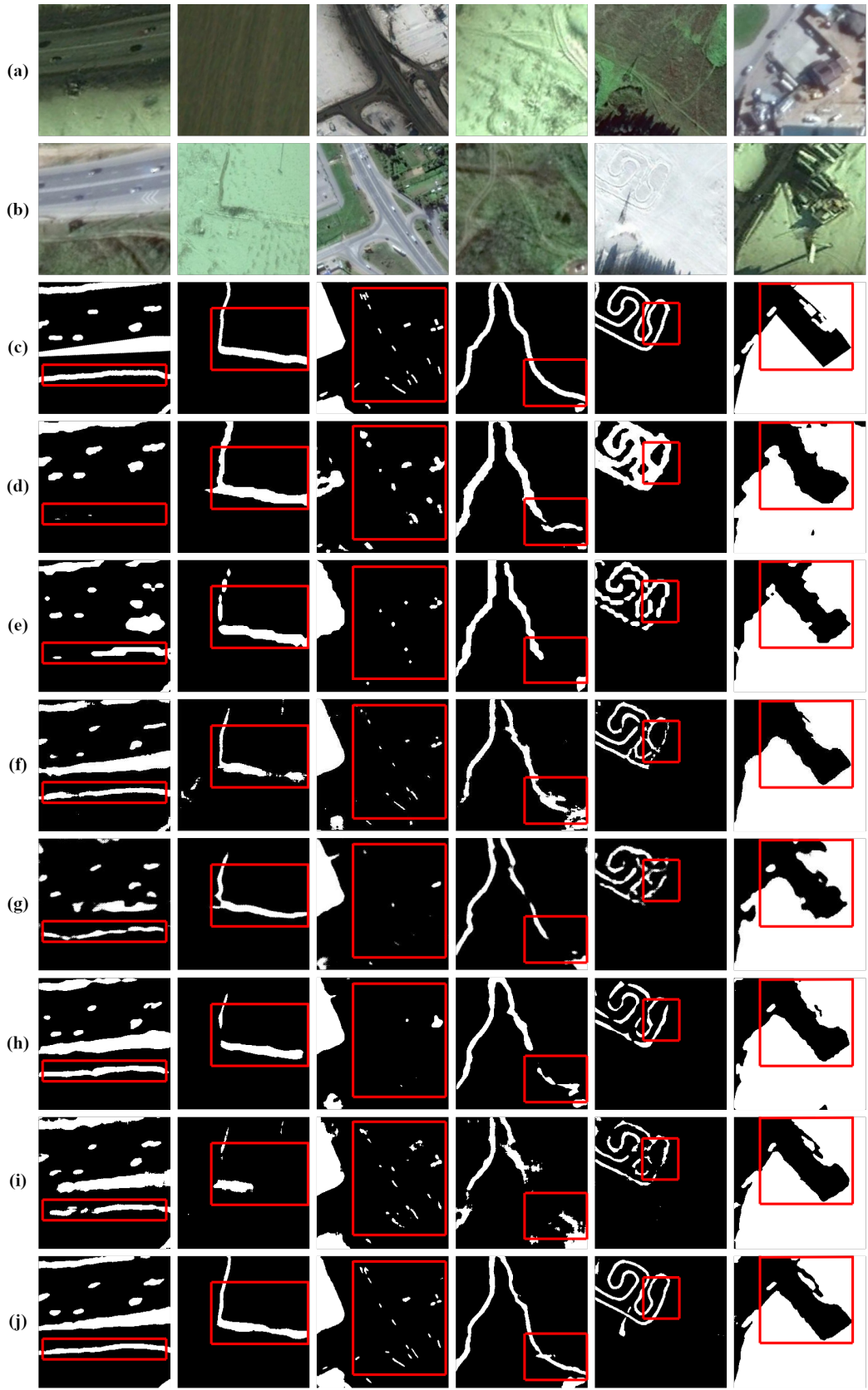

**Fig. 4** Visualized comparison results on CDD dataset. Images from top to bottom are (a) pre-event images, (b) post-event images, (c) ground truth, and the results of (d) STANet, (e) DASNet, (f) BIT, (g) FTN, (h) VcT, (i) SNUNet, and (j) our RFL-CDNet, respectively.

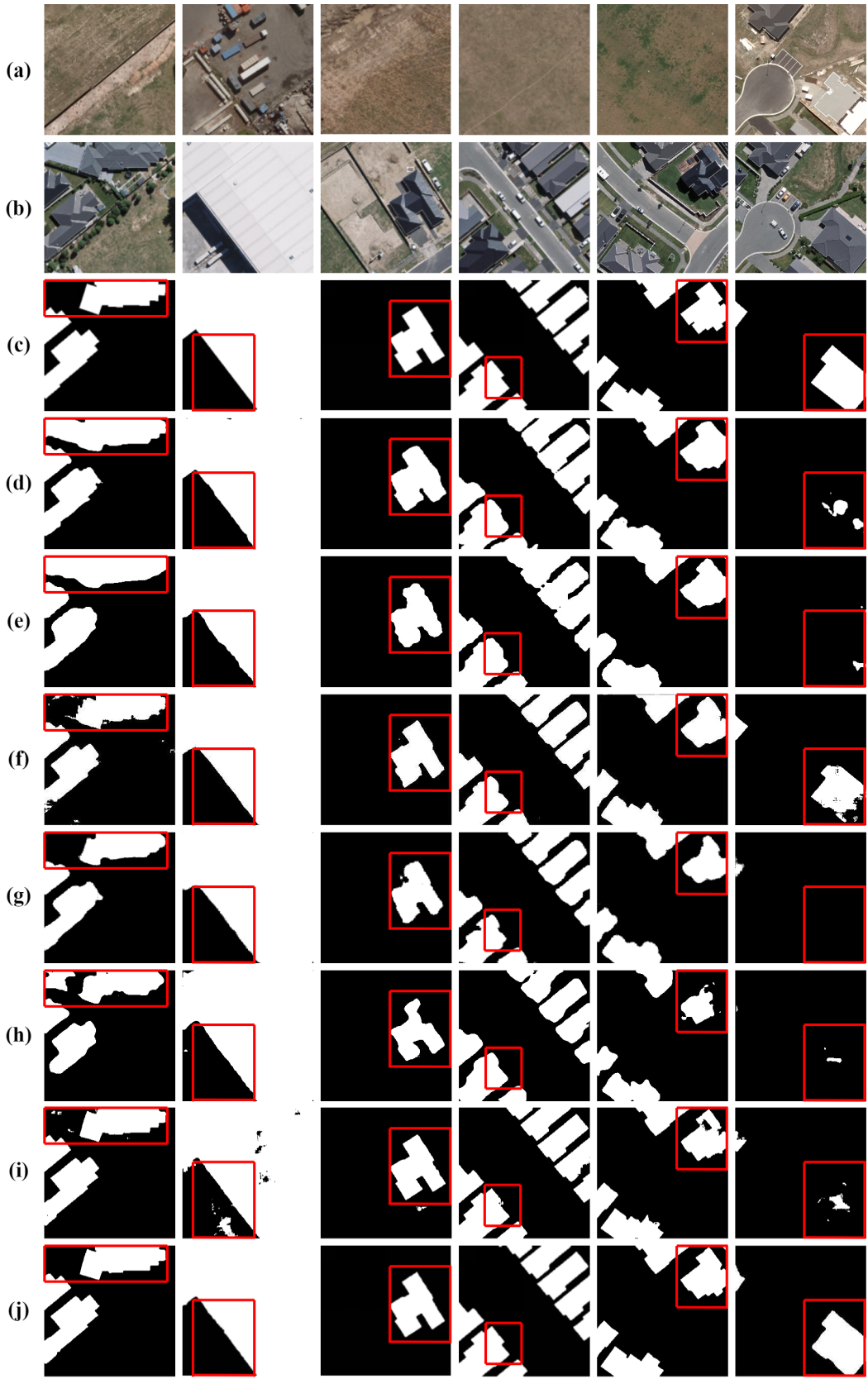

**Fig. 5** Visualized comparison results on WHU building dataset. Images from top to bottom are (a) pre-event images, (b) post-event images, (c) ground truth, and the results of (d) STANet, (e) DASNet, (f) BIT, (g) FTN, (h) VcT, (i) SNUNet, and (j) our RFL-CDNet, respectively.

*4.4 Ablation Studies*

In this section, we conduct ablation studies on WHU cultivated land dataset to evaluate the effectiveness of each component proposed in RFL-CDNet. In addition, we also reveal some insights into the framework design, *i.e.,* the influence of employing different stages to guide feature learning, the impact of different weighting strategies, and some discussions on the model size and computational complexity.

**Table 4.** Ablation studies of each module in our RFL-CDNet on WHU cultivated land dataset. "DMS" represents the model being trained under deep multiple supervision, "C2FG" indicates the model adopting a coarse-to-fine guidance strategy for richer feature learning, and "LF" means using our learnable fusion module to fuse multiple side predictions. The green arrows indicate the increase compared with their corresponding baseline.

| Methods | P (%) | R (%) | F1-score (%) |
|---|---|---|---|
| Baseline | 68.88 | 72.09 | 70.45 |
| + DMS | 69.27 | 72.94 | 71.06 (↑0.61) |
| + DMS+ C2FG | **71.19** | 72.38 | 71.78 (↑0.72) |
| + DMS+ C2FG + LF | 70.87 | **73.75** | **72.28** (↑0.50) |

*4.4.1 Effectiveness of Each Module*

We first verify the effectiveness of each module in our RFL-CDNet. Compared with the baseline SNUNet, we introduce deep multiple supervision, and propose C2FG and LF fusion modules respectively. We conducted ablation studies on WHU cultivated land dataset to evaluate how each module of RFL-CDNet affects the performance. The comparison experimental results are listed in Table 4. "Baseline" refers to SNUNet trained on WHU cultivated land dataset. "DMS" refers to the model trained with deep multiple supervision, while "C2FG" and "LF" are our designed coarse-to-fine guidance module and learnable fusion module, respectively. As shown in Table 4, all modules can effectively and consistently improve the performance. Specifically, deep multiple supervision can improve baseline module by a gain of 0.61% *F1-score.* Moreover, C2FG and LF modules can further improve performance of their corresponding baseline by the gain of 0.72% and 0.50% *F1-score*, respectively. From the perspective of performance improvement, the contribution of all modules is relatively equivalent. These experimental results prove that our improved design, including deep multiple supervision, C2FG and LF modules, all contribute to improve the performance of change detection.

*4.4.2 Influence of Each Stage in C2FG Module*

We further explore the influence of each stage in C2FG module for multi-scale representation learning. For ablation study, we gradually add LSTM cells to C2FG module without fusing side predictions. Table 5 lists the comparison results on WHU cultivated land dataset. The experimental setting involving only stage 0 for C2FG module is our baseline SNUNet, on which we gradually introduce more LSTM cells to C2FG. As shown in Table 5, the performance increases consistently as more stages are introduced, proving that each stage works well. Moreover, we also find that fine-scale stages lead to more performance gains. Among them, stage 1 can bring a gain of 0.82% *F1-score*, while the gains of stage 2 and stage 3 are only 0.27% and 0.24% *F1-score*, respectively. This is because the coarse-scale features capture semantics, and with more location information from fine scales, it enhances the boundaries and details of the detected change regions well. Overall, the performance gain in this study demonstrates the effectiveness of all stages and further verifies the effectiveness of our proposed C2FG module.

Table 5. Performance contribution of each stage in C2FG module.

| Stage ID | 0 | 1 | 2 | 3 | P (%) | R (%) | F1-score (%) |
|---|---|---|---|---|---|---|---|
| | √ | | | | 68.88 | 72.09 | 70.45 |
| C2FG Stage | √ | √ | | | 70.37 | 72.20 | 71.27 (↑0.82) |
| | √ | √ | √ | | 66.00 | 78.08 | 71.54 (↑0.27) |
| | √ | √ | √ | √ | 71.19 | 72.38 | 71.78 (↑0.24) |

*4.4.3 Influence of Different Weighting Strategies for LF*

We also compare three different fusion strategies for LF, including Global Weighted Fusion strategy (GWF), LF module with Maximum function (LF_Max), and LF module with Softmax function (LF_Softmax). GWF learns one scale-related weights for each side prediction via additional convolutional layers. While LF_Max employs the Maximum function to compute scale-spatial-related weighting parameters from the confidence map, and LF_Softmax adopts Softmax function instead. It should be noted that we adopt our initial model SNUNet equipped with C2FG module as our baseline in this study. Table 6 reports the experimental results. We notice that all fusion strategies yield better performance than the baseline, which proves the effectiveness of ensembling side predictions. Since LF module with softmax function performs best, we eventually adopt softmax function for weighting in our LF module.

Table 6. Ablation studies of different fusion strategies in LF module on WHU cultivated land dataset.

| Methods | P (%) | R (%) | F1-score (%) |
| --- | --- | --- | --- |
| Baseline | **71.19** | 72.38 | 71.78 |
| + GWF | 70.34 | 73.58 | 71.92 |
| + LF_Max | 69.78 | **74.25** | 71.95 |
| + LF_Softmax | 70.87 | 73.75 | **72.28** |

*4.4.3 Discussion of Model Size and Computational Complexity*

We further report the number of parameters and the computation cost in Table 7. Since we introduce deep multiple supervision, C2FG module and LF module to our framework, the efficiency of our RFL-CDNet decreases a little compared to the baseline SNUNet. Specifically, the parameters of our RFL-CDNet increase by merely 0.17M, and the increment of FLOPs is only 12.36G. Considering the performance improvement, *i.e.*, the gains of 1.83%, 4.87% and 2.17% *F1-score* on the WHU cultivated land, CDD and WHU building datasets respectively, the degree of efficiency drop is acceptable.

Table 7. Comparison of model size and computational complexity with previous methods.

| Methods | Params/M | FLOPs/G |
| --- | --- | --- |
| STANet | 16.93 | 26.33 |
| DASNet | 48.22 | 100.70 |
| SNUNet | 27.07 | 123.18 |
| Ours | 27.24 | 135.54 |

## 5  Conclusion

In this paper, we propose a novel framework for accurate change detection via exploiting rich intermediate features for multi-scale representation learning, termed RFL-CDNet. To fully leverage multi-scale features and enhance the capacity of feature learning at intermediate layers, we incorporate deep multiple supervision to the intermediate scales. Regarding rich intermediate features and side predictions, we design two modules: Coarse-To-Fine Guiding (C2FG) module and Learnable Fusion (LF) module. These modules serve to cultivate more discriminative multi-scale representations. Specifically, the C2FG module progressively aggregate semantics from coarse scales with the details from fine-scale features, while the LF module integrates side predictions according to the learned confidence map. We evaluate the effectiveness of our RFL-CDNet on three challenging datasets: WHU cultivated land, CDD, and WHU building datasets. Our proposed RFL-CDNet demonstrates remarkable performance, achieving an F1-score of 72.28% on WHU cultivated land dataset, 96.12% on CDD dataset, and 91.39% on WHU building dataset.

By far, we have demonstrated the effectiveness of our method on several change detection datasets. We highlight our C2FG and LF module also benefit other tasks and methods involving multi-scale representations, because we only assume that multi-scale representations are complementary in semantics and local details. For example, it is promising to apply C2FG and LF modules to dense prediction tasks like semantic segmentation, which would help to produce more accurate segmentation maps by incorporating semantics and detailed boundary structures from multi-scale features. However, the model size and computational cost of our method is relatively high because of the densely connected Nested-UNet backbone. In future works, we will explore compressing the backbone for higher efficiency through knowledge distillation, which is more friendly to real-world applications. Besides, it is also promising to incorporate the proposed C2FG and LF modules into other effective backbones like Transformers for improvements.

**Acknowledgments**

This work was supported in part by National Natural Science Foundation of China under Grant U23B2048, 62076186 and 62225113. The numerical calculations in this paper have been done on the supercomputing system in the Supercomputing Center of Wuhan University.